\documentclass[11pt]{article}

\usepackage{epsfig,amsmath,latexsym,amssymb}
\usepackage{graphicx}
\usepackage{lscape}
\usepackage{picture, eso-pic, tikz} 
\usepackage{dsfont}
\usepackage{listings}
\usepackage{changes}
\usepackage{hyperref}
\usepackage{bbding}
\usepackage{url}

\renewcommand{\baselinestretch}{1.1}
\def\R{{\mathbb R}}  
\def\p{{\mathbb P}}  
\def\E{{\mathbb E}}  %
\def\Beweis{\footnotesize}

\newcommand{\Remm}[1]{}
\newtheorem{theo}{Theorem}[section]
\newtheorem{lemma}[theo]{Lemma}
\newtheorem{prop}[theo]{Proposition}

\newtheorem{cor}[theo]{Corollary}

\newtheorem{model ass}[theo]{Model Assumptions}

\newtheorem{example}[theo]{Example}

\newtheorem{rems}[theo]{Remarks}

\def\EndProof{\hfill {\scriptsize $\Box$}}
\def\EndExample{\hfill {\scriptsize $\blacksquare$}}

\numberwithin{equation}{section}

\definecolor{MyGray}{rgb}{0.92,0.92,0.92}


\definecolor{British racing}{rgb}{0.0, 0.5, 0.0}

\def\bZ{\boldsymbol{Z}}

\def\bX{\boldsymbol{X}}
\def\b0{\boldsymbol{0}}

\def\b0{\boldsymbol{0}}

\usepackage{todonotes}
\newcommand{\Comments}{1}
\newcommand{\mynote}[2]{\ifnum\Comments=1\textcolor{#1}{#2}\fi}
\newcommand{\mytodo}[2]{\ifnum\Comments=1%
  \todo[linecolor=#1!80!black,backgroundcolor=#1,bordercolor=#1!80!black]{#2}\fi}

\ifnum\Comments=1               
\fi

\ifnum\Comments=1               
\fi

\ifnum\Comments=1               
\fi

\begin{document}
\author{Mario V.~W\"uthrich\footnote{RiskLab, Department of Mathematics, ETH Zurich,
    mario.wuethrich@math.ethz.ch}
}

\date{Version of \today}
\title{Model Selection with Gini Indices under Auto-Calibration
}
\maketitle

\begin{abstract}
\noindent  
The Gini index does not give a strictly consistent scoring rule in general. Therefore, maximizing the
Gini index may lead to wrong decisions. The main issue is that the Gini index is a rank-based
score that is not calibration-sensitive. We show that 
the Gini index allows for strictly consistent scoring if we restrict to the class of auto-calibrated 
regression models.

\bigskip

\end{abstract}

\section{Introduction}
The Gini index (Gini score, accuracy ratio) is a popular tool for model selection in machine learning, and
there are versions of the Gini index that are used 
to evaluate actuarial pricing models and financial credit risk models;
see Frees et al.~\cite{Frees1, Frees2}, Denuit et al.~\cite{DenuitSznajderTrufin}, Engelmann et al.~\cite{Engelmann}
and Tasche \cite{Tasche}.
However, in general, the Gini index does not give a (strictly) consistent scoring rule; 
Example 3 of Byrne \cite{Byrne} gives a counterexample. 
(Strict) consistency is an important property in model selection
because it ensures that maximizing the Gini index does not lead to a wrong
model choice;
see Gneiting \cite{Gneiting} and  Gneiting--Raftery \cite{GneitingRaftery}.
The Gini index can be obtained from
Somers' $D$ \cite{Somers}, which essentially considers Kendall's $\tau$; see
Newson \cite{Newson}. Intuitively, this tells us that the Gini index is a
rank-based score that is not calibration-sensitive.
The missing piece to make the 
Gini index a strictly consistent scoring rule is to restrict it to the class of 
auto-calibrated regression models, this is proved in Theorem \ref{proposition Gini}, below;
for auto-calibration we refer to Kr\"uger--Ziegel \cite{Ziegel},
Denuit et al.~\cite{DenuitCharpentierTrufin} and
Section 7.4.2 of W\"uthrich--Merz \cite{WM2022}. 

\medskip

{\bf Organization.} In the next section, we introduce the notion of strictly consistent scoring rules.
In Section \ref{The Gini index in machine learning}, we discuss the Gini index as it is usually
used in the machine learning community. In Section \ref{Auto-calibration and consistency of the Gini index},
we introduce and discuss the property of having an auto-calibrated regression model (forecasts),
and we prove that
the Gini index gives a strictly consistent scoring rule if we restrict to the class
of auto-calibrated regression models. This makes the maximization of the Gini index
a sensible model selection tool on the class of auto-calibrated
regression models. Finally, in Section \ref{Conclusions} we conclude.

\section{Consistent scoring rules}
Let $(Y,\bX)$ be a random tuple on a sufficiently rich
probability space $(\Omega, {\cal A}, \p)$ with real-valued
non-negative response $Y$ having finite mean
and with covariates $\bX$.
Denote by ${\cal F}$ the family of potential distributions of $(Y,\bX)$
being supported on ${\cal Y} \times {\cal X}$.
Let $F_{Y| \bX}$ be the conditional distribution of $Y$, given $\bX$.
For any model $(Y,\bX)\sim F \in {\cal F}$, we
consider the conditional mean functional $T$
\begin{equation*}
F_{Y|\bX} ~\mapsto~ T(F_{Y|\bX}) = \mu^\dagger(\bX)=
\E \left[\left.Y\right|\bX \right],
\end{equation*}
where
$\bX \mapsto \mu^\dagger(\bX)=\E \left[\left.Y\right|\bX \right]$ denotes the
true regression function of the chosen model.
The main task in regression modeling is to find this unknown true regression
function $\mu^\dagger(\cdot)$ from i.i.d.~data $(Y_i,\bX_i)$, $1\le i\le n$, having the same distribution
as $(Y,\bX)$.

\medskip

Choose a scoring function $S:{\cal Y}\times \R \to \R$
giving us the score $\E\left[S\left(Y,\widehat{\mu}(\bX)\right)\right]$ for 
regression function $\bX \mapsto \widehat{\mu}(\bX)$ and $(Y,\bX) \sim F \in {\cal F}$.
A {\it scoring rule} is obtained by selecting the argument(s) $\widehat{\mu}^\star(\cdot)$ that maximize
the score over the regression functions $\widehat{\mu}(\cdot)$, subject to existence,
\begin{equation}\label{scoring rule definition}
  \widehat{\mu}^\star(\cdot) ~\in~ \underset{\widehat{\mu}(\cdot)}{\arg\max}~
  \E\left[S\left(Y,\widehat{\mu}(\bX)\right)\right],
  \end{equation}
  under the given model choice $(Y,\bX) \sim F \in {\cal F}$.

  \medskip

A scoring
rule is called {\it consistent} on ${\cal F}$ for the conditional mean functional $T$,
if for any model $(Y,\bX)\sim F \in {\cal F}$ with conditional distributions 
$F_{Y|\bX}$ of $Y$, given $\bX$, 
we have $S(Y,T(F_{Y|{\bX}})) \in L^1(\p)$, and for any regression
function  $\bX \mapsto \widehat{\mu}(\bX)$ with $S(Y,\widehat{\mu}(\bX)) \in L^1(\p)$
we have
\begin{equation}\label{definition consistency}
\E\left[S\left(Y,T(F_{Y|{\bX}})\right)\right] \ge
\E\left[S\left(Y,\widehat{\mu}(\bX)\right)\right].
\end{equation}
A scoring rule is called {\it strictly consistent} on ${\cal F}$ for the conditional mean functional 
$T$, if it is consistent on ${\cal F}$,
and if an identity in \eqref{definition consistency} holds if and only if
$\widehat{\mu}(\bX)=T(F_{Y|\bX})=\mu^\dagger(\bX)$, a.s.

\begin{rems} \normalfont
  \begin{itemize}
  \item Strict consistency implies that the true regression function $\mu^\dagger(\cdot)$ is the
    unique maximizer in \eqref{scoring rule definition}, and it can be estimated by
    score maximization (assuming it is contained in the set over which we optimize, which we generally do).
    Empirically, we then consider for i.i.d.~data $(Y_i,\bX_i)$, $1\le i\le n$,
    \begin{equation*}
\underset{\widehat{\mu}(\cdot)}{\arg\max}~
  \frac{1}{n}\sum_{i=1}^nS\left(Y_i,\widehat{\mu}(\bX_i)\right),
  \end{equation*}
  where we still need to ensure that we can exchange the limit $n\to\infty$ and the $\arg\max$-operator to asymptotically select the true regression function $\mu^\dagger(\cdot)$ under strict consistency.
\item
  Formula \eqref{definition consistency} states unconditional consistency as we average over the
  distribution of $\bX$. 
For conditional consistency (in $\bX$) and its relation to the unconditional version we refer to Section 2.2 in 
Dimitriadis et al.~\cite{DimitriadisFisslerZiegel2020}. A point prediction
version of consistency is given in Definition 1 in Gneiting \cite{Gneiting}.
\item For scoring rule \eqref{scoring rule definition} we consider a maximization. By a sign switch we can
  turn this into a minimization problem, and in that case we rather speak about expected loss minimization.
\item Typically, we restrict \eqref{scoring rule definition}-\eqref{definition consistency} to smaller classes of regression
  functions $\bX \mapsto \widehat{\mu}(\bX)$. In the sequel, we will require continuity for these
  smaller classes, and, further below, we require the auto-calibration property. This requires
  that the true regression function $\mu^\dagger(\cdot)$ has this continuity, auto-calibration it will
  satisfy automatically, see Lemma \ref{lemma convex order}, below.
\end{itemize}
\end{rems}

\section{The Gini index in machine learning}
\label{The Gini index in machine learning}
In the sequel we 
assume $\widehat{\mu}(\bX)$ to have a continuous distribution $F_{\widehat{\mu}(\bX)}$ for all $(Y,\bX)\sim F \in {\cal F}$ and for any considered regression function $\bX \mapsto \widehat{\mu}(\bX)$.
This implies $F_{\widehat{\mu}(\bX)}(F_{\widehat{\mu}(\bX)}^{-1}(\alpha))=\alpha$ for all $\alpha \in (0,1)$, 
and with $F_{\widehat{\mu}(\bX)}^{-1}$
denoting the left-continuous generalized inverse of $F_{\widehat{\mu}(\bX)}$.

\medskip

In machine learning (ML) one considers the {\it cumulative accuracy profile} (CAP) defined by
\begin{equation*}
\alpha \in (0,1) \quad \mapsto \quad {\rm CAP}_{Y,\widehat{\mu}(\bX)} (\alpha) 
~=~ \frac{1}{\E[Y]}\, \E \left[ Y\,
\mathds{1}_{\left\{ \widehat{\mu}(\bX) > F_{\widehat{\mu}(\bX)}^{-1}(1-\alpha)\right\}}\right] ~\in~ [0,1].
\end{equation*}
In actuarial science, the CAP is also called concentration curve (up to sign switches), see Denuit--Trufin \cite{DenuitTrufin}.
The CAP measures a rank-based correlation between the prediction $\widehat{\mu}(\bX)$ and
the response $Y$. 

The {\it Gini index (Gini score, Gini ratio, Gini coefficient, accuracy ratio) in ML} 
is defined by
\begin{equation}\label{Gini ML}
G^{\rm ML}_{Y, \widehat{\mu}(\bX)} = \frac{\int_0^1 
{\rm CAP}_{Y,\widehat{\mu}(\bX)} (\alpha)\, d\alpha -1/2}
{\int_0^1 {\rm CAP}_{Y,Y} (\alpha)\, d\alpha -1/2},
\end{equation}
where we additionally assume that $Y$ has an (unconditional) continuous distribution $F_Y$.
For a geometric interpretation see Figure \ref{Gamma Lorenz} (lhs) and formula \eqref{Gini geometry}, below.

\begin{rems}\normalfont
  \begin{itemize}
  \item
The denominator in \eqref{Gini ML} does not use the regression function $\widehat{\mu}(\cdot)$, 
i.e., it has no impact on model selection by maximizing the Gini index $G^{\rm ML}_{Y, \widehat{\mu}(\bX)}$
over $\widehat{\mu}(\cdot)$. 
Hence, for scoring we can focus on the term in the
enumerator
\begin{eqnarray}\nonumber
\int_0^1 {\rm CAP}_{Y,\widehat{\mu}(\bX)} (\alpha)\, d\alpha
&=&\frac{1}{\E[Y]} \,
\E \left[ Y \int_0^1 
\mathds{1}_{\left\{ \widehat{\mu}(\bX) > F_{\widehat{\mu}(\bX)}^{-1}(1-\alpha)\right\}}\, d\alpha
    \right]
  \\&=&\nonumber
\frac{1}{\E[Y]} \,
        \E \left[ Y \,\p \left[\left.
        F_{\widehat{\mu}(\bX)}^{-1}(U)< \widehat{\mu}(\bX) \right| \widehat{\mu}(\bX)
    \right]        \right]
  \\&=&\frac{1}{\E[Y]} \,
        \E \left[ Y  F_{\widehat{\mu}(\bX)}(\widehat{\mu}(\bX)) \right],
\end{eqnarray}
for an independent $(0,1)$-uniform random variable $U$ and where we use continuity of
$F_{\widehat{\mu}(\bX)}$. This shows that the Gini index in ML is not calibration-sensitive
because $F_{\widehat{\mu}(\bX)}(\widehat{\mu}(\bX))$ has a $(0,1)$-uniform distribution, i.e.,
the specific distribution of $\widehat{\mu}(\bX)$ does not matter, but only its correlation
with $Y$ matters. 
\item Since typically the true data model $(Y,\bX) \sim F$ is not known,
  the Gini index in ML \eqref{Gini ML} is replaced by an empirical version
\begin{equation}\label{Gini ML empirical}
\widehat{G}^{\rm ML}_{Y, \widehat{\mu}(\bX)} = \frac{\int_0^1 
\widehat{\rm CAP}_{Y,\widehat{\mu}(\bX)} (\alpha)\, d\alpha -1/2}
{\int_0^1 \widehat{\rm CAP}_{Y,Y} (\alpha)\, d\alpha -1/2}~\le~1,
\end{equation}
where we set 
\begin{equation}\label{CAP empirical}
  \widehat{\rm CAP}_{Y,\widehat{\mu}(\bX)} (\alpha)
  =\frac{1}{\frac{1}{n}\sum_{i=1}^n Y_i}
\, \frac{1}{n} \sum_{i=1}^{n} Y_i\,
\mathds{1}_{\left\{\widehat{\mu}(\bX_i)> \widehat{\mu}\left(\bX_{\left(\lceil (1-\alpha) n \rceil\right)}\right)\right\}},
  \end{equation}
for i.i.d.~data $(Y_i,\bX_i)$, $1\le i\le n$, having the same distribution
as $(Y,\bX)$, and for order statistics 
$\widehat{\mu}(\bX_{(1)})< \widehat{\mu}(\bX_{(2)}) < \ldots <
\widehat{\mu}(\bX_{(n)})$;
note that by assumption the distribution of $\widehat{\mu}(\bX)$ is continuous which implies that
all observations $\widehat{\mu}(\bX_i)$ are mutually different for $1\le i \le n$, and we have a strict ordering
in the order statistics.
\item Let us further comment on \eqref{CAP empirical}. First, if we mirror the CAP at the diagonal we
have
\begin{eqnarray}\label{mirrored CAP definition}
{\rm CAP}^-_{Y,\widehat{\mu}(\bX)} (\alpha) 
&=& \frac{1}{\E[Y]}\, \E \left[ Y\,
\mathds{1}_{\left\{ \widehat{\mu}(\bX) \le F_{\widehat{\mu}(\bX)}^{-1}(\alpha)\right\}}\right]
  \\&=&\nonumber
        1-
\frac{1}{\E[Y]}\, \E \left[ Y\,
\mathds{1}_{\left\{ \widehat{\mu}(\bX) > F_{\widehat{\mu}(\bX)}^{-1}(\alpha)\right\}}\right]
~=~1-{\rm CAP}_{Y,\widehat{\mu}(\bX)} (1-\alpha). 
\end{eqnarray}
For an empirical version of the mirrored CAP we replace the above expression by
\begin{eqnarray*}
\widehat{\rm CAP}^-_{Y,\widehat{\mu}(\bX)} (\alpha) 
&=& \frac{1}{\frac{1}{n}\sum_{i=1}^n Y_i}\, 
\frac{1}{n} \sum_{i=1}^{n} Y_i\,
\mathds{1}_{\left\{\widehat{\mu}(\bX_i) \le \widehat{F}_{\widehat{\mu}(\bX)}^{-1}(\alpha)\right\}}
\\&=& 1-
\frac{1}{\frac{1}{n}\sum_{i=1}^n Y_i}\, 
\frac{1}{n} \sum_{i=1}^{n} Y_i\,
\mathds{1}_{\left\{\widehat{\mu}(\bX_i) > \widehat{\mu}(\bX_{(\lceil \alpha n\rceil)})\right\}},
\end{eqnarray*}
where in the last indicator we use the empirical distribution, for $m \in \R$ and $\alpha \in (0,1)$,
\begin{equation*}
\widehat{F}_{\widehat{\mu}(\bX)}(m)
= \frac{1}{n}\, \sum_{i=1}^n \mathds{1}_{\{ \widehat{\mu}(\bX_i) \le m \}}
\qquad \text{ and } \qquad
\widehat{F}^{-1}_{\widehat{\mu}(\bX)}(\alpha)
=  \widehat{\mu}(\bX_{(\lceil \alpha n\rceil)}).
\end{equation*}
This justifies the choice in \eqref{CAP empirical}. Similarly, we have
for the denominator in \eqref{Gini ML empirical}
\begin{equation}\label{discrete Lorenz curve continuous}
    \widehat{\rm CAP}^-_{Y,Y} (\alpha)=
\frac{1}{\frac{1}{n}\sum_{i=1}^n Y_i}
\, \frac{1}{n} \sum_{i=1}^{n} Y_i\,
\mathds{1}_{\left\{Y_i\,\le \,Y_{(\lceil \alpha n \rceil)}\right\}}
~\stackrel{(*)}{=}~
\frac{1}{\frac{1}{n}\sum_{i=1}^n Y_i}
\, \frac{1}{n} \sum_{i=1}^{\lceil \alpha n \rceil} Y_{(i)},
\end{equation}
for the identity $\stackrel{(*)}{=}$ to hold for any $\alpha \in (0,1)$, we need to assume that we have a strict ordering
$Y_{(1)}< Y_{(2)} < \ldots <Y_{(n)}$, i.e., that there are no ties in the
observations $(Y_i)_{1\le i \le n}$, which is the case because $Y$ was assumed to have a continuous
distribution $F_Y$.
This then motivates to set
\begin{eqnarray}\nonumber
  \widehat{\rm CAP}_{Y,Y} (\alpha) &=& 1-\widehat{\rm CAP}^-_{Y,Y} (1-\alpha)
                                       =
\frac{1}{\frac{1}{n}\sum_{i=1}^n Y_i}
\, \frac{1}{n} \sum_{i=1}^{n} Y_i\, \mathds{1}_{\left\{Y_i\,> \,Y_{(\lceil (1-\alpha) n \rceil)}\right\}}
\\&\stackrel{(*)}{=}&
                                       \frac{1}{\frac{1}{n}\sum_{i=1}^n Y_i}
\, \frac{1}{n} \sum_{i=\lceil (1-\alpha) n \rceil+1}^n Y_{(i)}
.
\label{Lorenz empirical}
\end{eqnarray}
If we have a perfect joint ordering  between
$(Y_i)_{1\le i \le n}$ and $(\widehat{\mu}(\bX_i))_{1\le i \le n}$, the upper bound in \eqref{Gini ML empirical} is attained,
see \eqref{CAP empirical} and \eqref{Lorenz empirical}. This is the motivation for the scaling in
\eqref{Gini ML}.
\end{itemize}
\end{rems}

In the definition of the Gini index in ML \eqref{Gini ML} we have assumed that $Y$ has a continuous
distribution $F_Y$. This is not the case for discrete responses $Y$. Therefore, in the discrete case
we need to replace the denominator in \eqref{Gini ML} by a different object. For illustrative
purposes we show the binary classification case in the next example.

\begin{example}[binary classification]\normalfont
  We consider a binary classification example with true regression function
  \begin{equation*}
    \bX~\mapsto~ p^\dagger(\bX)= \E \left[\left. Y \right| \bX \right]=    \p \left[\left. Y=1 \right| \bX \right]
    ~\in ~(0,1).
    \end{equation*}
    That is, $Y$ is conditionally Bernoulli distributed, given $\bX$, with probability
    $p^\dagger(\bX) \in (0,1)$ and range ${\cal Y}=\{0,1\}$. In this case the CAP
    for a regression function $\bX \mapsto \widehat{p}(\bX)$
    with continuous distribution $F_{\widehat{p}(\bX)}$ is for $\alpha \in (0,1)$ given by 
    \begin{eqnarray*}
      {\rm CAP}_{Y,\widehat{p}(\bX)} (\alpha) 
&=& \frac{1}{\E[Y]}\, \E \left[ Y\,
    \mathds{1}_{\left\{ \widehat{p}(\bX) > F_{\widehat{p}(\bX)}^{-1}(1-\alpha)\right\}}\right]
     \\&=& \frac{1}{\p[Y=1]}\, \E \left[ \mathds{1}_{\left\{Y=1, \,
    \widehat{p}(\bX) > F_{\widehat{p}(\bX)}^{-1}(1-\alpha)\right\}}\right]
      \\&=&\p \left[\left. \widehat{p}(\bX) > F_{\widehat{p}(\bX)}^{-1}(1-\alpha)\right| Y=1\right]
      \\&=&1-F_{\widehat{p}(\bX)|Y=1}\left(F_{\widehat{p}(\bX)}^{-1}(1-\alpha)\right).
\end{eqnarray*}
This corresponds to formula (5.2) in Tasche \cite{Tasche}.

For the Gini index in ML we need to calculate the
denominator of \eqref{Gini ML}. However, this formula only applies
for a continuous distribution $F_Y$ of $Y$.
In the case of a discrete distribution of $Y$ we need to modify \eqref{Gini ML}.
Starting from the right-hand side of \eqref{discrete Lorenz curve continuous}, we define the empirical function
in the discrete case by
\begin{equation*}
  \alpha\in(0,1) ~ \mapsto ~  \widehat{\rm CAP}^-_{Y,Y} (\alpha) ~=~
\frac{1}{\frac{1}{n}\sum_{i=1}^n Y_i}
\, \frac{1}{n} \sum_{i=1}^{\lceil \alpha n \rceil} Y_{(i)},
\end{equation*}
for i.i.d.~data $(Y_i,\bX_i)$, $1\le i\le n$.
In the Bernoulli case, this function is identically equal to zero up to $\alpha \le 1-\sum_{i=1}^n Y_i/n$, these
describes the number of zeros among the observations $(Y_i)_{1\le i \le n}$, and afterwards it increases
to 1. Since this increase is only described on the discrete grid with span $1/n$, we linearly
interpolate between these points. This provides a straight line between
$1-\sum_{i=1}^n Y_i/n$ and 1 with slope $n/\sum_{i=1}^n Y_i$. Under this linear interpolation,
we get the area (integral)
\begin{equation*}
  \int_0^1 \widehat{\rm CAP}^-_{Y,Y} (\alpha) \, d\alpha = \frac{1}{2n}\,\sum_{i=1}^n Y_i.
\end{equation*}  
By the law of large numbers, the latter converges to $p^\dagger/2=\E[p^\dagger(\bX)]/2=\E[Y]/2$, a.s.,
as $n\to \infty$.
This motivates in the (discrete) binary classification case the following
definition of the Gini index in ML
\begin{equation}\label{Gini ML Bernoulli}
G^{\rm ML}_{Y, \widehat{\mu}(\bX)} = \frac{1/2-\int_0^1 
F_{\widehat{p}(\bX)|Y=1}\left(F_{\widehat{p}(\bX)}^{-1}(1-\alpha)\right) d\alpha}
{(1-p^\dagger)/2}.
\end{equation}
In the binary classification case, the CAP can be related to the receiver operating characteristics (ROC)
curve. The area under the curve (AUC) of the ROC curve has a one-to-one relationship to the
Gini index in ML \eqref{Gini ML Bernoulli} in the Bernoulli case, we refer to Section 5 in Tasche
\cite{Tasche}. We mention this because the ML community more frequently uses the AUC than the
Gini index for model selection.

In general, in the discrete case we replace the integral in the denominator in \eqref{Gini ML}
by the term
\begin{equation}\label{the general denominator}
\frac{1}{4\E[Y]}\,
\E\left[\left|Y-\widetilde{Y}\right|\right],
\end{equation}
where $\widetilde{Y}$ is an independent copy of $Y$. 
This latter quantity \eqref{the general denominator}
can be calculated for any distribution $F_Y$ of $Y$, and in the continuous case we
precisely receive the denominator in \eqref{Gini ML}. The binary classification case
\eqref{the general denominator} provides us with $(1-p^\dagger)/2$ which gives
\eqref{Gini ML Bernoulli}.
\EndExample
\end{example}

\section{Auto-calibration and consistency of the Gini index}
\label{Auto-calibration and consistency of the Gini index}
Let $(Y,\bX) \sim F$.
A regression function $\bX \mapsto \widehat{\mu}(\bX)$ is
auto-calibrated for $Y$ if, a.s., 
\begin{equation*}
\widehat{\mu}(\bX) = \E \left[ Y \left|\widehat{\mu}(\bX) \right]\right. .
\end{equation*}
Auto-calibration is an important property in insurance pricing, as it implies that every
cohort of insurance policies paying the same price $\widehat{\mu}(\bX)$
is in average self-financing, because the price $\widehat{\mu}(\bX)$ exactly
covers the expected claim $Y$ of that cohort. I.e., we do not 
have any systematic cross-financing between the price cohorts.
This is the core of risk classification in insurance.
It also implies
unbiasedness on the portfolio level
\begin{equation}\label{unbiasedness under auto-calibration}
\E\left[\widehat{\mu}(\bX)\right] = \E \left[ Y \right],
\end{equation} 
which is  a minimal requirement in insurance pricing. Typically, there
are many auto-calibrated regression functions $\widehat{\mu}(\bX)$ for $Y$, i.e., there are many
systems of self-financing pricing cohorts. 

\begin{lemma}
\label{lemma convex order}
The true regression function $\bX\mapsto\mu^\dagger(\bX)=\E[Y|\bX ]$ is auto-calibrated for $Y$,
and it strictly dominates in convex order
any other auto-calibrated regression function $\bX\mapsto \widehat{\mu}(\bX)$ for $Y$.
\end{lemma}

{\Beweis
{\bf Proof.} To prove auto-calibration of $\mu^\dagger$ we apply the tower property
to the $\sigma$-algebras $\sigma( \mu^\dagger(\bX))
\subset \sigma(\bX)$ which gives, a.s.,
\begin{equation*}
\E\left[Y\left| \mu^\dagger(\bX)\right]\right.
=\E\left[\E\left[Y\left| \bX\right]\right.\left| \mu^\dagger(\bX)\right]\right.
=\E\left[\mu^\dagger(\bX)\left| \mu^\dagger(\bX)\right]\right.
=\mu^\dagger(\bX).
\end{equation*}
For any convex function $\psi$, auto-calibration, the
tower property for $\sigma(\widehat{\mu}(\bX))
\subset \sigma(\bX)$ and
Jensen's inequality give
\begin{eqnarray*}
\E \left[ \psi \left(\widehat{\mu}(\bX) \right)\right]
&=&\E \left[ \psi \left(\E \left[ Y \left|\widehat{\mu}(\bX) \right]\right.\right)\right]
~=~\E \left[ \psi \left(\E \left[\E \left[ Y \left|\bX \right]\right.\left|\widehat{\mu}(\bX) \right]\right.\right)\right]
~=~\E \left[ \psi \left(\E \left[\left.\mu^\dagger(\bX)\right|\widehat{\mu}(\bX) \right]\right)\right]
\\&\le &\E \left[ \E \left[\left.\psi\left(\mu^\dagger(\bX)\right) \right|\widehat{\mu}(\bX) \right]\right]
~=~ \E \left[  \psi\left(\mu^\dagger(\bX)\right) \right],
\end{eqnarray*}
whenever these exist. This proves that $\mu^\dagger$ dominates in convex order any other auto-calibrated
regression function $\widehat{\mu}$ for $Y$.
Assume that there exists an auto-calibrated regression function
$\bX\mapsto \widehat{\mu}(\bX)$ for $Y$ such that for any convex function $\psi$
we have an equality in the previous calculation, whenever these exist.
This implies that $\mu^\dagger(\bX)$ is $\sigma(\widehat{\mu}(\bX))$-measurable.
Auto-calibration and the tower property for $\sigma(\widehat{\mu}(\bX))
\subset \sigma(\bX)$ then
provide, a.s., 
\begin{equation*}
  \widehat{\mu}(\bX) = \E \left[ Y \left|\widehat{\mu}(\bX) \right]\right.
  = \E \left[\E \left[ Y \left|\bX \right]\right. \left|\widehat{\mu}(\bX) \right]\right.
  =\E \left[\left. \mu^\dagger(\bX) \right|\widehat{\mu}(\bX) \right]=\mu^\dagger(\bX).
\end{equation*}
This proves the statement of strict convex order.
\EndProof}

\medskip

The next proposition is a consequence of Lemma \ref{lemma convex order}
and of Theorem 3.1 in Kr\"uger--Ziegel \cite{Ziegel}.

\begin{prop} \label{proposition Bregman}
The true regression function $\bX\mapsto\mu^\dagger(\bX)$ forecast-dominates any
auto-calibrated regression function $\bX \mapsto \widehat{\mu}(\bX)$ for $Y$ meaning that
\begin{equation*}
\E \left[ - D_{\psi}(Y,\mu^\dagger(\bX))\right] \ge \E \Big[ -D_{\psi}(Y,\widehat{\mu}(\bX))\Big],
\end{equation*}
for any convex function $\psi$ where the above exists, and with Bregman divergence
given by
\begin{equation*}
D_\psi(y,m) = \psi(y)-\psi(m)-\psi'(m)(y-m) ~\ge ~0,
\end{equation*}
for $y,m \in \R$ and $\psi'$ is a (sub-)gradient of the convex function $\psi$.
\end{prop}

Proposition \ref{proposition Bregman} says that every negative Bregman divergence provides a consistent
scoring rule \eqref{definition consistency} for the conditional mean regression functional $T$ 
under auto-calibration for $Y$. 
This statement motivates the common practice 
in model selection of minimizing (out-of-sample) deviance
losses, as deviance losses are special cases of Bregman divergences; see
Chapters 2 and 4 in W\"uthrich--Merz \cite{WM2022}. For more information on this topic
we refer to Kr\"uger--Ziegel \cite{Ziegel}, Theorem 7 in Gneiting \cite{Gneiting} 
and Savage \cite{Savage}, the latter two references state that Bregman divergences provide
the only strictly consistent scoring functions for mean estimation.

\medskip

The definition of the Gini index \cite{Gini0} in economics slightly differs from the
ML version \eqref{Gini ML}. Assume $F_{\widehat{\mu}(\bX)}$ is a continuous distribution.
It is then based on the
{\it Lorenz curve} \cite{Lorenz} given by
\begin{equation*}
\alpha \in (0,1) \quad \mapsto \quad L_{{\widehat{\mu}(\bX)}} \left( F_{\widehat{\mu}(\bX)}^{-1}(\alpha) \right) 
~=~ \frac{1}{\E[\widehat{\mu}(\bX)]}\, \E \left[ \widehat{\mu}(\bX)
\mathds{1}_{\left\{ \widehat{\mu}(\bX) \le F_{\widehat{\mu}(\bX)}^{-1}(\alpha)\right\}}\right] ~\in~ [0,1].
\end{equation*}
Note that we have the property ${\rm CAP}^-_{\widehat{\mu}(\bX), \widehat{\mu}(\bX)}=
L_{{\widehat{\mu}(\bX)}} ( F_{\widehat{\mu}(\bX)}^{-1}(\alpha))$, see \eqref{mirrored CAP definition}.

\medskip

The {\it Gini index in economics} has many (equivalent)\footnote{For an equivalence
  in \eqref{Gini economics} we need that $F_{\widehat{\mu}(\bX)}$ is continuous,
  otherwise one should choose the term on the right-hand side as the definition
  of the Gini index in economics.}
  definitions, we use the following two
\begin{equation}\label{Gini economics}
G^{\rm eco}_{\widehat{\mu}(\bX)} = 1 - 2 \int_0^1 L_{{\widehat{\mu}(\bX)}} \left( F_{\widehat{\mu}(\bX)}^{-1}(\alpha)\right) d\alpha
=
\frac{1}{2\E[\widehat{\mu}(\bX)]}\,
\E\Big[\Big|\widehat{\mu}(\bX)-\widehat{\mu}(\bZ)\Big|\Big],
\end{equation}
where $\widehat{\mu}(\bZ)$ is an independent copy of $\widehat{\mu}(\bX)$. The first definition
in \eqref{Gini economics} is based on a continuous distribution  $F_{\widehat{\mu}(\bX)}$, whereas
the second one can be used for any distribution  $F_{\widehat{\mu}(\bX)}$, we also refer to \eqref{the general denominator}.

\begin{figure}[htb!]
\begin{center}
\begin{minipage}[t]{0.44\textwidth}
\begin{center}
\includegraphics[width=\textwidth]{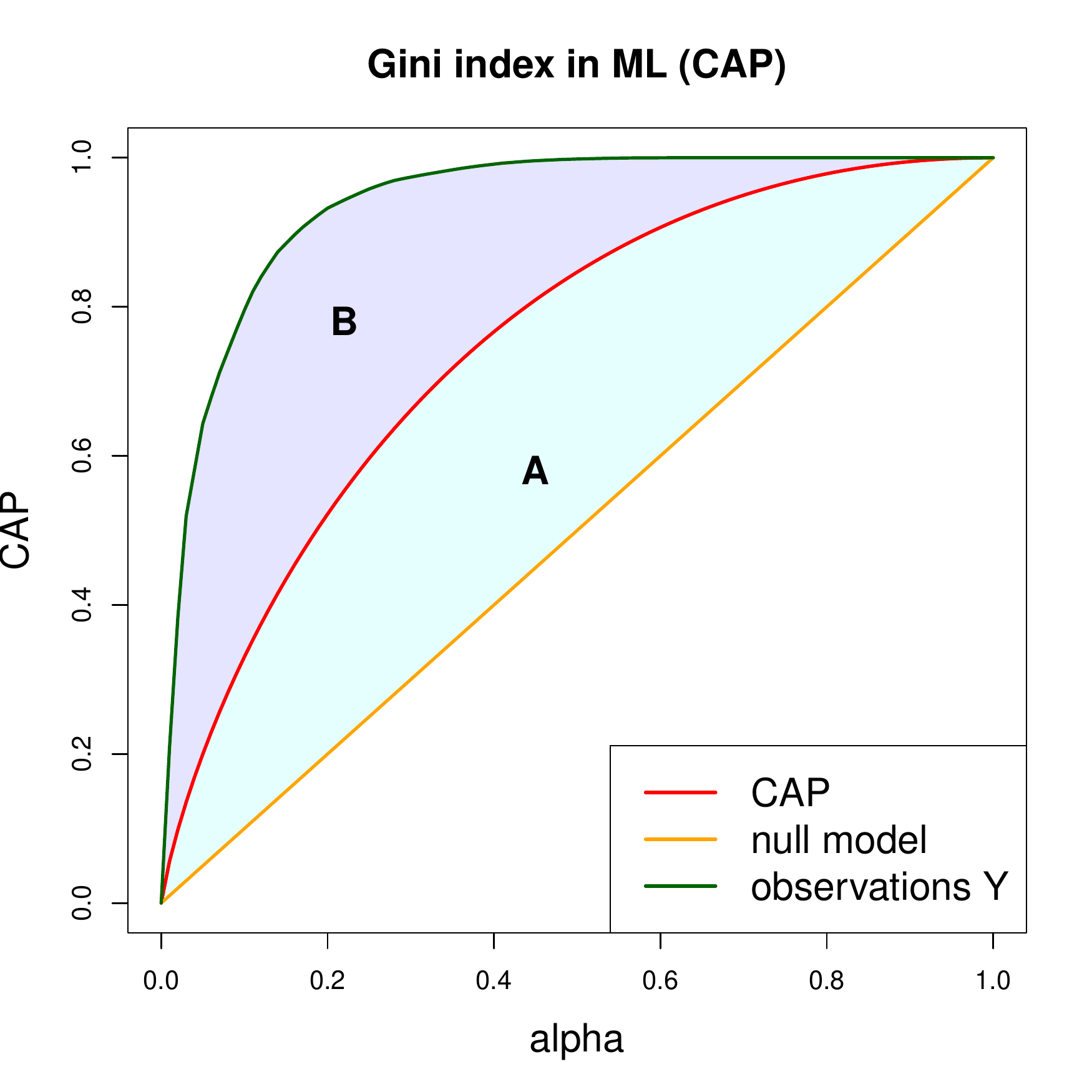}
\end{center}
\end{minipage}
\begin{minipage}[t]{0.44\textwidth}
\begin{center}
\includegraphics[width=\textwidth]{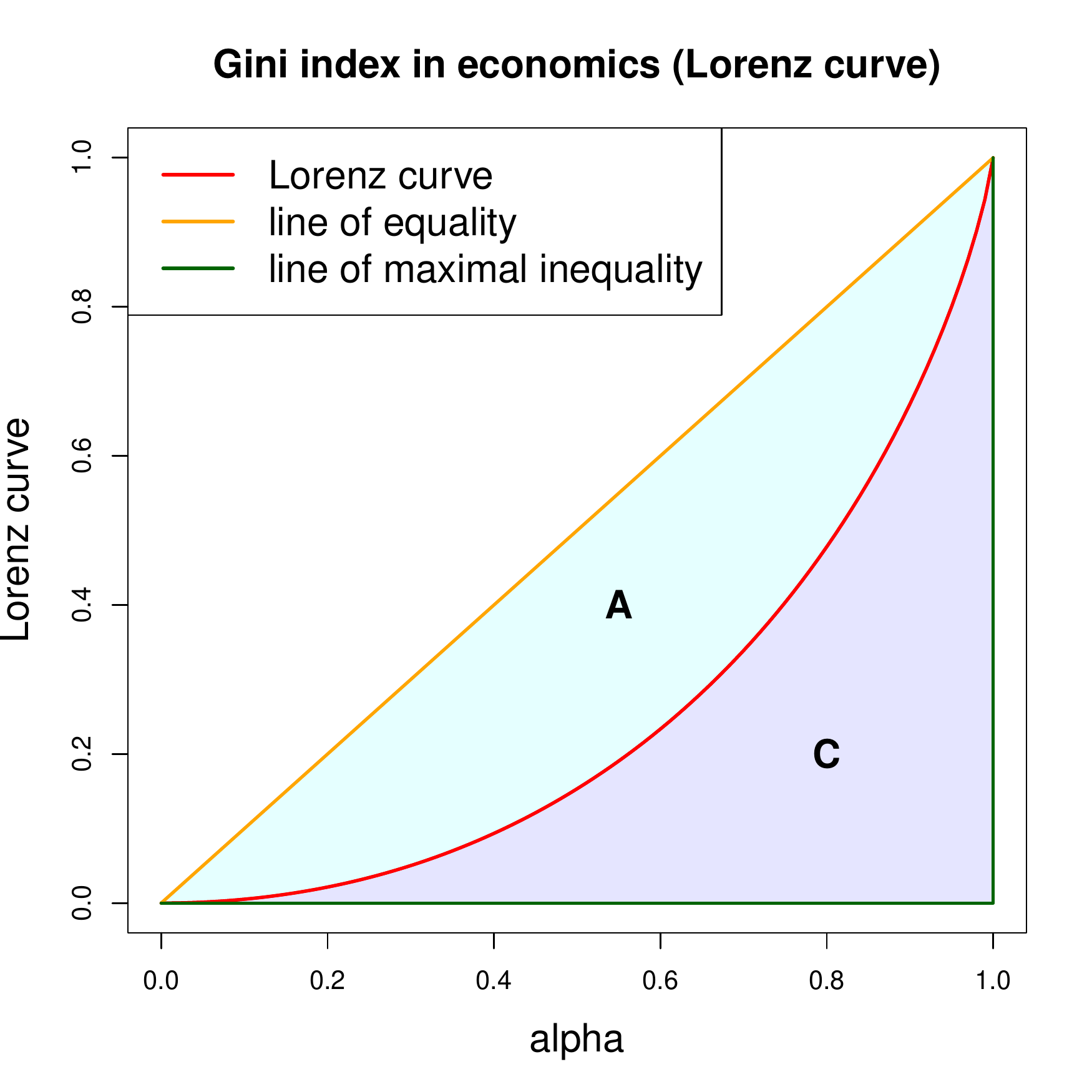}
\end{center}
\end{minipage}
\end{center}
\vspace{-.7cm}
\caption{(lhs) cumulative accuracy profile (CAP) and (rhs) Lorenz curve.}
\label{Gamma Lorenz}
\end{figure}

There are three differences between the Gini index in ML and the one in economics, 
see Figure \ref{Gamma Lorenz}:
(i) $G^{\rm eco}_{\widehat{\mu}(\bX)}$ 
considers a mirrored version of the curves compared to $G^{\rm ML}_{Y, \widehat{\mu}(\bX)}$;
(ii) $G^{\rm ML}_{Y, \widehat{\mu}(\bX)}$ depends on $Y$ and $\widehat{\mu}(\bX)$, 
$G^{\rm eco}_{\widehat{\mu}(\bX)}$ only depends on $\widehat{\mu}(\bX)$; (iii) scalings are different leading to areas
B and C, respectively, in Figure \ref{Gamma Lorenz}. The two Gini indices are 
geometrically obtained by, see Figure \ref{Gamma Lorenz},
\begin{equation}\label{Gini geometry}
G^{\rm ML}_{Y,\widehat{\mu}(\bX)} = 
\frac{\text{area}({\rm A})}{\text{area}({\rm A}+{\rm B})} 
\quad \text{ and } \quad G^{\rm eco}_{\widehat{\mu}(\bX)} = 
\frac{\text{area}({\rm A})}{\text{area}({\rm A}+{\rm C})}= 2\,
\text{area}({\rm A})=1-2\,\text{area}({\rm C}).
\end{equation}

Property 3.1 of Denuit--Trufin \cite{DenuitTrufin} gives the following nice result.
\begin{prop}\label{Denuit Trufin result}
Under auto-calibration of the regression function $\bX \to \widehat{\mu}(\bX)$ for $Y$ we have 
the identity ${\rm CAP}_{Y,\widehat{\mu}(\bX)} (\alpha)=1-L_{\widehat{\mu}(\bX)}
( F_{\widehat{\mu}(\bX)}^{-1}(1-\alpha))$ for 
all $\alpha \in (0,1)$.
\end{prop}

{\Beweis
  {\bf Proof.} Using the tower property, auto-calibration of $\widehat{\mu}$ for $Y$
  and unbiasedness \eqref{unbiasedness under auto-calibration} give us
\begin{eqnarray*}
{\rm CAP}_{Y,\widehat{\mu}(\bX)} (\alpha) 
&=& \frac{1}{\E[Y]}\, \E \left[ Y\,
\mathds{1}_{\left\{ \widehat{\mu}(\bX) > F_{\widehat{\mu}(\bX)}^{-1}(1-\alpha)\right\}}\right]
~=~ \frac{1}{\E[Y]}\, \E \left[\E \left[\left. Y \right|  \widehat{\mu}(\bX)\right]\,
\mathds{1}_{\left\{ \widehat{\mu}(\bX) > F_{\widehat{\mu}(\bX)}^{-1}(1-\alpha)\right\}}\right]
\\
&=& \frac{1}{\E[\widehat{\mu}(\bX)]}\, \E \left[  \widehat{\mu}(\bX)\,
\mathds{1}_{\left\{ \widehat{\mu}(\bX) > F_{\widehat{\mu}(\bX)}^{-1}(1-\alpha)\right\}}\right]
~=~1-\frac{1}{\E[\widehat{\mu}(\bX)]}\, \E \left[  \widehat{\mu}(\bX)
\mathds{1}_{\left\{ \widehat{\mu}(\bX) \le F_{\widehat{\mu}(\bX)}^{-1}(1-\alpha)\right\}}\right].
\end{eqnarray*}
This proves the claim.
\EndProof}

\medskip

Thus, under auto-calibration for $Y$, the CAP and the Lorenz curve coincide (up to mirroring/sign
switching).
This gives us the following corollary.
\begin{cor} \label{identical Gini}
  Under auto-calibration of the regression function $\bX \mapsto \widehat{\mu}(\bX)$ for $Y$ we have
  for the Gini indices
\begin{equation}\label{GL identity}
G^{\rm ML}_{Y,\widehat{\mu}(\bX)}=
\frac{G^{\rm eco}_{\widehat{\mu}(\bX)}}
{2\int_0^1 
{\rm CAP}_{Y,Y} (\alpha)\, d\alpha -1}.
\end{equation}
\end{cor}

{\Beweis
  {\bf Proof.} Proposition \ref{Denuit Trufin result} gives us for the Gini index in ML
\begin{eqnarray*}
G^{\rm ML}_{Y, \widehat{\mu}(\bX)} &=& \frac{\int_0^1 
{\rm CAP}_{Y,\widehat{\mu}(\bX)} (\alpha)\, d\alpha -1/2}
{\int_0^1 {\rm CAP}_{Y,Y} (\alpha)\, d\alpha -1/2}
~=~
\frac{1/2-\int_0^1 
L_{\widehat{\mu}(\bX)}
( F_{\widehat{\mu}(\bX)}^{-1}(1-\alpha))\, d\alpha}
{\int_0^1 {\rm CAP}_{Y,Y} (\alpha)\, d\alpha -1/2}
\\&=&
\frac{1-2\int_0^1 
L_{\widehat{\mu}(\bX)}
( F_{\widehat{\mu}(\bX)}^{-1}(1-\alpha))\, d\alpha}
{2\int_0^1 {\rm CAP}_{Y,Y} (\alpha)\, d\alpha -1}
~=~
\frac{1-2\int_0^1 
L_{\widehat{\mu}(\bX)}
( F_{\widehat{\mu}(\bX)}^{-1}(\beta))\, d\beta}
{2\int_0^1 {\rm CAP}_{Y,Y} (\alpha)\, d\alpha -1},
\end{eqnarray*}
where the last step uses the change of variable $\alpha \mapsto \beta=1-\alpha$.
This proves the claim.
\EndProof}

\medskip

This says that under auto-calibration for the response both Gini indices (the ML score and the version in economics) provide the same scoring rule because the (positive) denominator\footnote{Note
  that the denominator in \eqref{GL identity} is positive for every non-deterministic $Y$.
  This follows from the fact that the denominator is equal to twice \eqref{the general denominator} which
  is positive unless $Y$ is deterministic.} in \eqref{GL identity} does not depend on 
the specific choice of the regression function $\widehat{\mu}(\cdot)$.
  Moreover, the same arguments
apply to the Gini indices in non-continuous cases, e.g., in the binary classification (Bernoulli)
case \eqref{Gini ML Bernoulli}.

\begin{theo} \label{proposition Gini}
The true regression function $\bX\mapsto\mu^\dagger(\bX)$ maximizes the Gini index (in ML)
among all auto-calibrated regression functions $\bX\mapsto\widehat{\mu}(\bX)$ for $Y$, i.e., 
$G^{\rm ML}_{Y,\mu^\dagger(\bX)} > G^{\rm ML}_{Y,\widehat{\mu}(\bX)}$ unless $\widehat{\mu}(\bX)=\mu^\dagger(\bX)$, a.s.
\end{theo}

{\Beweis
{\bf Proof.} Conditionally, given $\widehat{\mu}(\bZ)$, $m \mapsto |m-\widehat{\mu}(\bZ)|$ is a convex
function in $m \in \R$. Using  formula \eqref{Gini economics}, independence between $\widehat{\mu}(\bX)$ and $\widehat{\mu}(\bZ)$
in \eqref{Gini economics} and
Lemma \ref{lemma convex order} we obtain inequality, a.s.,
\begin{equation*}
\E\left[\left.\left|\widehat{\mu}(\bX)-\widehat{\mu}(\bZ)\right|\, \right| \widehat{\mu}(\bZ) \right]
\le 
\E\left[\left.\left|\mu^\dagger(\bX)-\widehat{\mu}(\bZ)\right|\, \right| \widehat{\mu}(\bZ) \right],
\end{equation*}
where $\mu^\dagger(\bX)$ is independent of $\widehat{\mu}(\bZ)$.
Using the tower property, 
applying the same argument to the exchanged role of $\mu^\dagger(\bX)$ and $\widehat{\mu}(\bZ)$,
using unbiasedness \eqref{unbiasedness under auto-calibration}
and using Corollary \ref{identical Gini}
provides $G^{\rm ML}_{Y,\mu^\dagger(\bX)} \ge G^{\rm ML}_{Y,\widehat{\mu}(\bX)}$.

Assume there exists an auto-calibrated regression function $\widehat{\mu}$ for $Y$ such that
$G^{\rm ML}_{Y,\mu^\dagger(\bX)} = G^{\rm ML}_{Y,\widehat{\mu}(\bX)}$. Using 
auto-calibration of $\widehat{\mu}$ for $Y$ and the tower property,
we receive for $\p$-a.e.~$\omega \in \Omega$
\begin{equation}\label{only using auto-calibration}
\widehat{\mu}(\bX)(\omega)
      =\E\left[Y \left|\widehat{\mu}(\bX)\right]\right.(\omega)
      =\E\left[\E\left.\left[Y\right|\bX\right] \left|\widehat{\mu}(\bX)\right]\right.(\omega)
      =\E\left[\left.\mu^\dagger(\bX) \right|\widehat{\mu}(\bX)\right](\omega).
    \end{equation}
    Denote by $\Omega_1 \subset \Omega$ a set of full measure 1 on which \eqref{only using auto-calibration}
    holds. On $\Omega_1$, the predictor 
$\widehat{\mu}(\bX)$ is between the conditional essential infimum and supremum of $\mu^\dagger(\bX)$,
given $\widehat{\mu}(\bX)$, because
it corresponds to the conditional expectation of $\mu^\dagger(\bX)$, given $\widehat{\mu}(\bX)$.
Consider the case of sample points
$\omega \in \Omega_1$ where the conditional essential infimum and supremum of $\mu^\dagger(\bX)$,
given $\widehat{\mu}(\bX)$, do not coincide, and denote the corresponding set
of sample points by $\Omega_2 \subset \Omega_1$. On $\Omega_2$,  
the predictor $\widehat{\mu}(\bX)$ is strictly between the conditional essential infimum and supremum of 
$\mu^\dagger(\bX)$, given $\widehat{\mu}(\bX)$, due to the
conditional expectation property \eqref{only using auto-calibration}.
We have using  \eqref{only using auto-calibration}
and independence between $\widehat{\mu}(\bX)$ and $\widehat{\mu}(\bZ)$
\begin{eqnarray}\nonumber
   \E\left[\left|\widehat{\mu}(\bX) -\widehat{\mu}(\bZ)\right| \right]
      &=&
  \E\left[\left|\E\left[\left.\mu^\dagger(\bX) \right|\widehat{\mu}(\bX)\right]-\widehat{\mu}(\bZ)\right| \right]
      ~=~\E\left[
          \left|\E\left[\left.\mu^\dagger(\bX) -\widehat{\mu}(\bZ) \right|\widehat{\mu}(\bX),\widehat{\mu}(\bZ)\right]\right| \right]
  \\&=&
        \E\left[\left(\mathds{1}_{\Omega_2}+\mathds{1}_{\Omega_2^c}\right)
        \left|\E\left[\left.\mu^\dagger(\bX) -\widehat{\mu}(\bZ) \right|\widehat{\mu}(\bX),\widehat{\mu}(\bZ)\right]\right| \right].
\label{this will give the proof}
\end{eqnarray}
We calculate the first term on the right-hand side of \eqref{this will give the proof}
\begin{equation*}
\E\left[\mathds{1}_{\Omega_2}
        \left|\E\left[\left.\mu^\dagger(\bX) -\widehat{\mu}(\bZ) \right|\widehat{\mu}(\bX),\widehat{\mu}(\bZ)\right]
          \right|\,\right]
=  \int_{\Omega_2}\left(
  \int_\Omega
 \left|\E\left[\left.\mu^\dagger(\bX) -\widehat{\mu}(\bZ) \right|\widehat{\mu}(\bX),\widehat{\mu}(\bZ)\right]\right|
      (\omega, \widetilde{\omega})\,d\p(\widetilde{\omega})\right) d\p(\omega).        
\end{equation*}
We study the inner integral for fixed sample point $\omega \in \Omega_2$. Jensen's inequality
gives us
\begin{equation}\label{Omega1}
 \int_\Omega
\left|\E\left[\left.\mu^\dagger(\bX) -\widehat{\mu}(\bZ) \right|\widehat{\mu}(\bX),\widehat{\mu}(\bZ)\right]\right|
     (\omega, \widetilde{\omega})\,d\p(\widetilde{\omega})
~<~  \int_\Omega   
\E\left[\left.\left|\mu^\dagger(\bX) -\widehat{\mu}(\bZ)\right|\, \right|\widehat{\mu}(\bX),\widehat{\mu}(\bZ)\right]
(\omega, \widetilde{\omega})\,d\p(\widetilde{\omega}),
\end{equation}
where we receive a strict inequality for $\omega \in \Omega_2$ because of
the following items: (1) on $\Omega_2$,
$\mu^\dagger(\bX)$ is non-deterministic, conditionally given $\widehat{\mu}(\bX)$, 
(2) $m \mapsto |m-\widehat{\mu}(\bZ)|$ is a convex function, 
(3) $\widehat{\mu}(\bZ)$ has the same distribution (and support) as $\widehat{\mu}(\bX)$,
and (4) $\widehat{\mu}(\bZ)$ and $(\mu^\dagger(\bX),\widehat{\mu}(\bX))$ are independent.
Items (1)-(4) imply that on a set of positive $\p(\widetilde{\omega})$-measure we receive
a strict Jensen's inequality, because on this set, $\widehat{\mu}(\bZ)$ is strictly within the conditional
essential infimum and supremum of (the non-deterministic) $\mu^\dagger(\bX)$, given $\widehat{\mu}(\bX)$.

Assume $\p[\Omega_2]>0$, i.e., strict inequality \eqref{Omega1} occurs on a set of positive measure.
Applying Jensen's inequality also to the other term in \eqref{this will give the proof}
we receive strict inequality
\begin{equation*}\nonumber
   \E\left[\left|\widehat{\mu}(\bX) -\widehat{\mu}(\bZ)\right| \right]
  ~< ~      \E\left[
    \E\left[\left.\left|\mu^\dagger(\bX) -\widehat{\mu}(\bZ)\right| \right|\widehat{\mu}(\bX),\widehat{\mu}(\bZ)\right] \right]
  =       \E\left[
        \left|\mu^\dagger(\bX) -\widehat{\mu}(\bZ)\right| \right].
\end{equation*}
This strict inequality contradicts our assumption $G^{\rm ML}_{Y,\mu^\dagger(\bX)} = G^{\rm ML}_{Y,\widehat{\mu}(\bX)}$.
Therefore, $\p[\Omega_2]=0$, which implies
\begin{equation*}
  \p \left[\Omega_2^c \cap \Omega_1\right] = \p \left[ \Omega_2^c \right]=1.
  \end{equation*}
On the set $\Omega_2^c \cap \Omega_1$, we have $\mu^\dagger(\bX) =\widehat{\mu}(\bX)$, which proves the claim.
\EndProof}

\medskip

Theorem \ref{proposition Gini} proves that the Gini index gives a strictly consistent
scoring rule on the class of auto-calibrated regression functions
that are $\bX$-measurable, because the true regression function 
$\bX\mapsto\mu^\dagger(\bX)$ maximizes this Gini index. A bigger Gini index can only
be achieved by a larger information set than the $\sigma$-algebra generated by $\bX$.

\medskip

The following proposition generalizes Property 5.1 of Denuit et al.~\cite{DenuitCharpentierTrufin},
which gives a method
of restoring auto-calibration for a general regression function $\bX \mapsto \widehat{\mu}(\bX)$.

\begin{prop}Consider a regression function $\bX \mapsto \widehat{\mu}(\bX)$. The following
  regression function is auto-calibrated for $Y$
  \begin{equation*}
    \bX~\mapsto~ \widehat{\mu}^{\rm (auto)}(\bX)= \E \left[Y \left| \widehat{\mu}(\bX) \right]\right. .
  \end{equation*}  

\end{prop}

{\Beweis
  {\bf Proof.} Note that $\widehat{\mu}^{\rm (auto)}(\bX)$ is $\sigma(\widehat{\mu}(\bX))$-measurable.
  This implies $\sigma(\widehat{\mu}^{\rm (auto)}(\bX)) \subset \sigma(\widehat{\mu}(\bX))$. Henceforth,
  using the tower property,  a.s.,
  \begin{equation*}
    \E \left[Y \left| \widehat{\mu}^{\rm (auto)}(\bX) \right]\right.
    = \E \left[\E \left[Y \left| \widehat{\mu}(\bX) \right]\right. \left| \widehat{\mu}^{\rm (auto)}(\bX) \right]\right.
    = \E \left[\widehat{\mu}^{\rm (auto)}(\bX) \left| \widehat{\mu}^{\rm (auto)}(\bX) \right]\right.
    =\widehat{\mu}^{\rm (auto)}(\bX).
  \end{equation*}
  This completes the proof.
  \EndProof}

\section{Conclusions}
\label{Conclusions}
In general, one should not use the Gini index for model selection because it
does not give a strictly consistent scoring rule and, thus, may lead to wrong decisions. We have shown
in Theorem \ref{proposition Gini}
that if we restrict Gini index scoring to the class of auto-calibrated regression functions for the
given response,
the Gini index allows for strictly consistent scoring. This also translates to the binary classification case where the
(machine learning version of the) Gini index has an equivalent formulation in terms of the
area under the curve (AUC) of the receiver operating characteristics (ROC) curve, we refer to
Tasche \cite{Tasche}.
We only need to ensure that the binary classification model is auto-calibrated for the
Bernoulli response to receive a strictly consistent scoring rule from the AUC.

\bigskip

{\small 
\renewcommand{\baselinestretch}{.51}
}

\end{document}